\documentclass[final]{cvpr}
\usepackage{float}
\usepackage{times}
\usepackage{epsfig}
\usepackage{graphicx}
\usepackage{amsmath}
\usepackage{amssymb}
\usepackage{bbm}
\usepackage{multirow}
\usepackage{color}
\usepackage{bm}
\usepackage{paralist}

\usepackage[pagebackref=false,breaklinks=true,colorlinks,bookmarks=false]{hyperref}

\def\cvprPaperID{11068} 
\def\confYear{CVPR 2021}

\begin{document}

\title{Positive Sample Propagation along the Audio-Visual Event Line}

\author{Jinxing Zhou$^{1,2}$\quad Liang Zheng$^{3*}$\quad Yiran Zhong$^3$\quad Shijie Hao$^{1,2}$\quad Meng Wang$^{1,2}$\thanks{Corresponding author.}\\
$^1$Hefei University of Technology\\
$^2$Intelligent Interconnected Systems Laboratory of Anhui Province\\
$^3$Australian National University\\
{\tt\small \{zhoujxhfut, hfut.hsj, eric.mengwang\}@gmail.com,}\\
{\tt\small \{liang.zheng, yiran.zhong\}@anu.edu.au}
}

\maketitle

\begin{abstract}
   Visual and audio signals often coexist in natural environments, forming audio-visual events (AVEs).
   Given a video, we aim to localize video segments containing an AVE
   and identify its category.
   In order to learn discriminative features for a classifier, it is pivotal to identify the helpful (or positive) audio-visual segment pairs while filtering out the irrelevant ones, regardless whether they are synchronized or not. 
   To this end, we propose a new positive sample propagation (PSP) module to discover and exploit the closely related audio-visual pairs
   by evaluating the relationship within every possible pair. 
   It can be done by constructing an all-pair similarity map between each audio and visual segment, and only aggregating the features from the pairs with high similarity scores. To encourage the network to extract high correlated features for positive samples, a new audio-visual pair similarity loss is proposed. We also propose a new weighting branch to better exploit the temporal correlations in weakly supervised setting. We perform extensive experiments on the public AVE dataset and achieve new state-of-the-art accuracy in both fully and weakly supervised settings, thus verifying the effectiveness of our method.

   
\end{abstract}

\vspace{-3mm}
\section{Introduction}\label{introduction}
\vspace{-0.1cm}

Recent literature has shown that by fusing multi-modality information can lead to better deep feature presentation, \ie, audio-visual fusion~\cite{arandjelovic2017look} and text-visual fusion~\cite{radford2021learning}. However, building a large scale multi-modality pre-training datasets would require heavy manual labours to clean and annotate the raw video sets. To relief the manual labour, recent work either focuses on learning from noise supervision~\cite{cheng2019noise, jia2021scaling} or tries to automatically filter out unpaired samples~\cite{tian2018audio}. 
\begin{figure}[t]
   \begin{center}
   \includegraphics[width=0.45\textwidth]{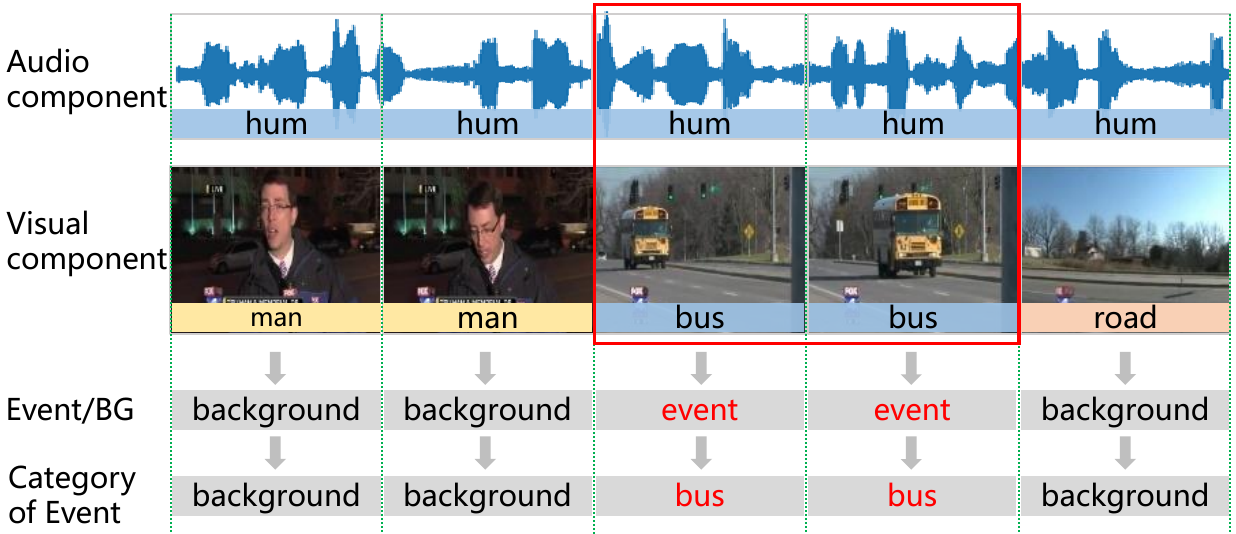}
   \end{center}
   \vspace{-0.4cm}
      \caption{An illustration of the AVE localization task.
      Each video segment is composed of an audio and a visual component.
      In this example,
      the ``hum'' of the bus exists in all the segments (audio modality), but
      the visual images of the ``bus'' only appear in the third and fourth segments (visual modality).
      So only these two segments (red boxes) are localized as (bus) {\em event}, the remaining are recognized as {\em background}.
      }
   \label{figure_1}
\end{figure}

The task of Audio-Visual Event (AVE) localization~\cite{tian2018audio} is served for the latter purpose. An AVE often refers as an event that is both audible and visible in a video segment, \ie, a sound source appears in an image ({\em visible}) while the source of the sound also exists in audio portion ({\em audible}). As shown in Fig.~\ref{figure_1}, a bus humming is an AVE in the third and fourth segments as we can see a bus and hear it humming simultaneously in these video segments. The AVE localization task is to find these video segments that contain an audio-visual event and classify it into a certain category\footnote{Note that there is a fundamental difference between the Multimedia Event Detection (MED) task and the AVE localization task: MED is a retrieval task that aims to find video clips that are associated with a particular event from a video archive while AVE localization is a classification problem.}. 

There are two relations that need to be considered in the AVE task: intra-modal relations and cross-modal relations. The former often addresses temporal relations in one single modality while the later also takes audio and visual relations into account. The pioneer work~\cite{lin2019dual, tian2018audio} often tries to regress the class by concatenating features from synchronized audio-visual pairs. Since these methods do not explicitly consider the intra-modal or cross-modal relations, their accuracy is often unsatisfying. The following works~\cite{tian2020han, wu2019dual, xu2020MM, xuan2020cross} utilize a self-attention mechanism to explicitly encode the temporal relations within intra-modality and some of them~\cite{ramaswamy2020makes, ramaswamy2020see, xu2020MM, xuan2020cross} also aggregate better audio-visual feature representations by encoding cross-modal relations. However, these methods often ignore the interference caused by irrelevant audio-visual segment pairs during the fusion process. In this paper, we argue that by only aggregating features from positive samples, \ie, high-relevant audio-visual pairs, we can have better AVE localization accuracy.

Specifically, we propose a new Positive Sample Propagation (PSP) module. In a nutshell, PSP first constructs an all-pair similarity map between each audio and visual segment and cuts off the entries that are below a pre-set similarity threshold, and then aggregates the audio and visual features without considering the negative and weak entries in an online fashion. Through various visualizations we show that PSP allows more relevant features that are not necessarily synchronized to be aggregated in an online fashion. 

Apart from PSP that can be used in both fully and weakly supervised settings (refer Sec.~\ref{problemdefine} for the setting details), we further propose two improvements that work under each setting, respectively. On the one hand, an audio-visual pair similarity loss is introduced under the fully supervised setting that encourages the network to learn high correlated features of audio and visual segments if they belong to the same event. On the other hand, we propose a weighting branch in the weakly supervised setting, which gives temporal weights to the segment features.

We evaluate our method on the standard AVE dataset~\cite{tian2018audio}. We show that the proposed techniques consistently benefit our system and when combined allow us to achieve state-of-the-art performance under both fully and weakly supervised settings.

\vspace{-0.1cm}
\section{Related work}
\vspace{-0.1cm}

\textbf{Audio-visual correspondence (AVC)} aims to predict whether a given visual image corresponds
duration of the audio.
A model is asked to judge whether the audio and visual signals describe the same object,
\eg, dog {\em v.s.} bark, cat {\em v.s.} meow.
It is a self-supervised problem since the visual image is usually accompanied by the corresponding sound.
Existing methods
try to evaluate the correspondences by measuring the audio-visual similarity~\cite{arandjelovic2017look, arandjelovic2018objects, aytar2016soundnet, cheng2020look, fayek2020large, hu2019deep}.
It will get a large similarity score if the audio-visual pair is corresponding, otherwise, a low score.
This motivates us to tackle the abundant audio-visual pairs in the AVE localization problem by considering the audio-visual similarity.

\textbf{Sound source localization} aims to localize those visual regions which are relevant to the provided audio signal.
It is related to {\em sound source separation}~\cite{afouras20ssl, darrell2000audio, parekh2017motion, pu2017audio, zhao2018sound} problem.
The target region of the visual frame must be corresponding with the given sound.
It is similar to the AVC task from this point of view,
but the real challenge of sound source localization
is to accurately locate the sound-maker when there are multiple sound sources in a visual frame.
Qian \etal~\cite{qian2020multiple}
adapt the Grad-CAM~\cite{selvaraju2017grad} to disentangle class-specific features for multiple sound sources problem.
Senocak \etal~\cite{senocak2018learning} propose a triplet loss working in an unsupervised manner.
Afouras \etal~\cite{afouras20ssl} utilize
a contrastive loss to train the model in a self-supervised learning way.
Both of these methods~\cite{afouras20ssl, senocak2018learning}
need to construct positive and negative audio-visual pair samples.
Since similar positive and negative samples are easily obtained in AVE localization, depending on whether the audio and visual segments depict the same event, 
we try to research those audio-visual pairs and explore its effect.

\textbf{Audio-visual event localization} aims to distinguish those segments including an
audio-visual event from a long video.
Existing works mainly focus on the audio-visual fusion process.
A dual multimodal residual network is proposed in~\cite{tian2018audio}.
Lin \etal~\cite{lin2019dual} adapt a bi-directional LSTM~\cite{schuster1997bidirectional} to fuse audio and visual features in a seq2seq manner.
During the whole fusion process, simple concatenation and addition operations
are adapted along the single synchronized audio-visual pair.
Ramaswamy~\cite{ramaswamy2020makes, ramaswamy2020see} utilizes a bilinear method to capture cross-modal relations.
Xuan \etal~\cite{xuan2020cross} propose to leverage {\em modality sentinel} to
give different weights to audio and visual features.
Lin \etal~\cite{Lin_2020_ACCV} design an audio-visual transformer to describe local spatial and temporal information.
The visual frame is divided into patches and adjacent frames are utilized, making the model complicated and computationally intensive.
Xu \etal~\cite{xu2020MM} attempt to leverage concatenating audio-visual features as the supervision
then the feature of each modality is updated by separate modules.
Unlike these, the proposed PSP method has a further in-depth study on the abundant audio-visual pairs,
selecting the most relevant ones.
Relying on these positive samples, more distinguished audio-visual features can be obtained after feature aggregation.

\begin{figure*}[t]
   \begin{center}
      \setlength{\abovecaptionskip}{0.cm}
      \includegraphics[width=\textwidth]{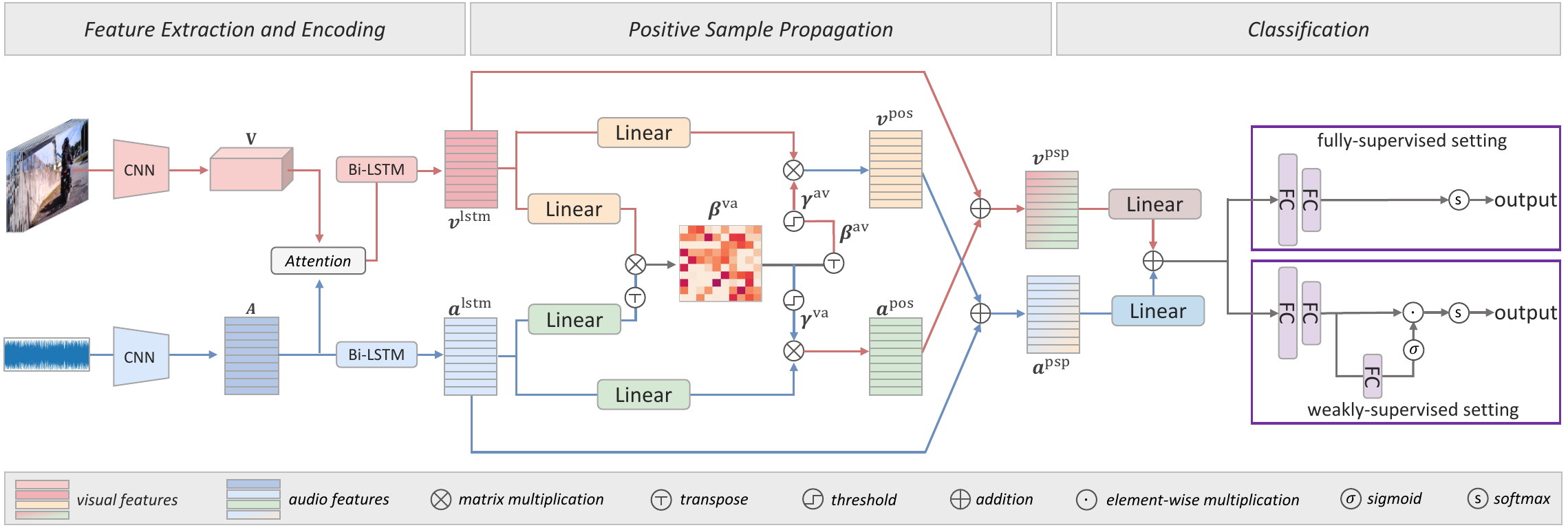}
   \end{center}
   \vspace{-0.4cm}
      \caption{System Flow. We first extract and encode video and audio features through existing modules such as AVGA~\cite{tian2018audio} and Bi-LSTM.
      The proposed positive sample propagation (PSP) takes the LSTM encoded features as input, which are fed to a few linear layers. An affinity matrix is computed before selecting the positive connections of audio-visual segment pairs using thresholding. In this module, 
      audio and visual features are aggregated by feature propagation through the positive connections. In the last stage, we classify the event into predefined categories. For the supervised setting, apart from the commonly used CE loss, we further propose an audio-visual pair similarity loss which enforces similar features between them when they contain an event. For the weakly supervised setting, we introduce another FC layer that gives weights to different video segments: higher weights are given event-containing segments.
      }
   \label{figure_3}
  \vspace{-0.6cm}
\end{figure*}
   
\vspace{-0.1cm}
\section{Problem statement}
\label{problemdefine}
\vspace{-0.1cm}
AVE localization aims to find out those segments
containing an audio-visual event \cite{tian2018audio}.
In other words, AVE localization is expected to decide whether each synchronized audio-visual pair depicts an event.
Besides, AVE localization needs to identify the event category for each segment. 
Specifically, a video sequence $S$ is divided into $T$ non-overlapping yet continuous segments $\{S_t^v, S_t^a\}^{T}_{t=1}$, and each segment is one-second long. $S^v$ and $S^a$ are the visual and audio components, respectively. We consider two settings of this task, to be described below.

\textbf{Fully-supervised AVE localization.}
Under the fully-supervised setting, the event label of every video segment is given, indicating
whether the segment denotes an event and which category the event belongs to.
We denote the event label of the $t^{th}$ segment as $\bm{y}_t = { \{ y_t^c | y_t^c \in \{0, 1\},  \sum_{c=1}^{C} y_t^c = 1\} \in \mathbb{R}^C}$,
where $C$ is the number of categories (including the {\em background}).
Then, the label for the entire video can be written as $\bm{Y}^{\text{fully}} = [\bm{y}_1; \bm{y}_2; ...; \bm{y}_T] \in \mathbb{R}^{T \times C}$. Through $\bm{Y}^{\text{fully}}$, we know whether an arbitrary synchronized audio-visual pair at time $t$ is an event: if the $1$ of its event label $\bm{y}_t$ is at the entry of a certain event instead of the \emph{background},
the pair describes an event, and otherwise does not.

\textbf{Weakly-supervised AVE localization.}
We adapt the weakly-supervised setting proposed in \cite{lin2019dual,xuan2020cross}, where the label $\bm{Y}^{\text{weak}} \in \mathbb{R}^{1 \times C}$ is the average pooling value of $\bm{Y}^{\text{fully}}$ along the column. It
implies the proportion of audio-visual pairs that contain an event. 
This setting is different from the fully-supervised one because the event label of each segment $y_t$ is unknown, making the problem more challenging. 

\vspace{-0.1cm}
\section{Our method}

\subsection{Overall pipeline}
\vspace{-0.1cm}
The overall pipeline of our system is illustrated in Fig.~\ref{figure_3},
which includes three modules:
a feature extraction and encoding module,
a positive sample propagation module, and a classification module.
In the {\em feature extraction and encoding} module,
audio-guided visual attention (AVGA~\cite{tian2018audio}) is adapted for early fusion to make the model focus on those visual regions closely related to the audio component.
Then a Bi-LSTM is utilized to encode temporal relations in video segments.
The LSTM encoded features are sent to the proposed
{\em positive sample propagation (PSP)} module.
PSP is able to select those positive connections of audio-visual segment pairs by measuring the cross-modal similarity with thresholding.
Audio and visual features are aggregated by feature propagation through the positive connections.
The updated audio-visual features after PSP are fused then sent to the final {\em classification} module, predicting which video segments contain an event and the event category.

\subsection{Feature extraction and encoding}\label{sec:feature_extractor}
\vspace{-0.1cm}

The visual and synchronized audio segments are processed by pretrained convolutional neural networks (CNNs).
We denote the resulting visual feature as $ \pmb V \in \mathbb{R}^{T \times N \times d_v } $,
where $d_v$ is the feature dimension, $N = H \times W$, $H$ and $W$ are the height and width of the feature map, respectively.
The extracted audio feature is denoted as $ \pmb A \in \mathbb{R}^{T \times d_a} $, where $d_a$ denotes feature dimension.
We then directly adapt AGVA~\cite{tian2018audio} for multi-modal early fusion. AVGA allows the model to focus on visual regions that are relevant to the audio component.
To encode the temporal relationship in video sequences, the visual and
audio features after AVGA are further sent to two independent Bi-LSTMs.
The updated visual and audio features are
represented as $\bm{v}^{\text{lstm}}\in \mathbb{R}^{T \times d_l}$ and $\bm{a}^{\text{lstm}} \in \mathbb{R}^{T \times d_l}$, respectively.

\subsection{Positive sample propagation (PSP)}\label{sec:PSP}
\vspace{-0.1cm}
PSP allows the network to learn more representative features
by exploiting the similarities of audio-visual pairs.
It involves three steps. 

\begin{figure}[t]
   \begin{center}
   \setlength{\abovecaptionskip}{0.cm}
   \includegraphics[width=0.45\textwidth]{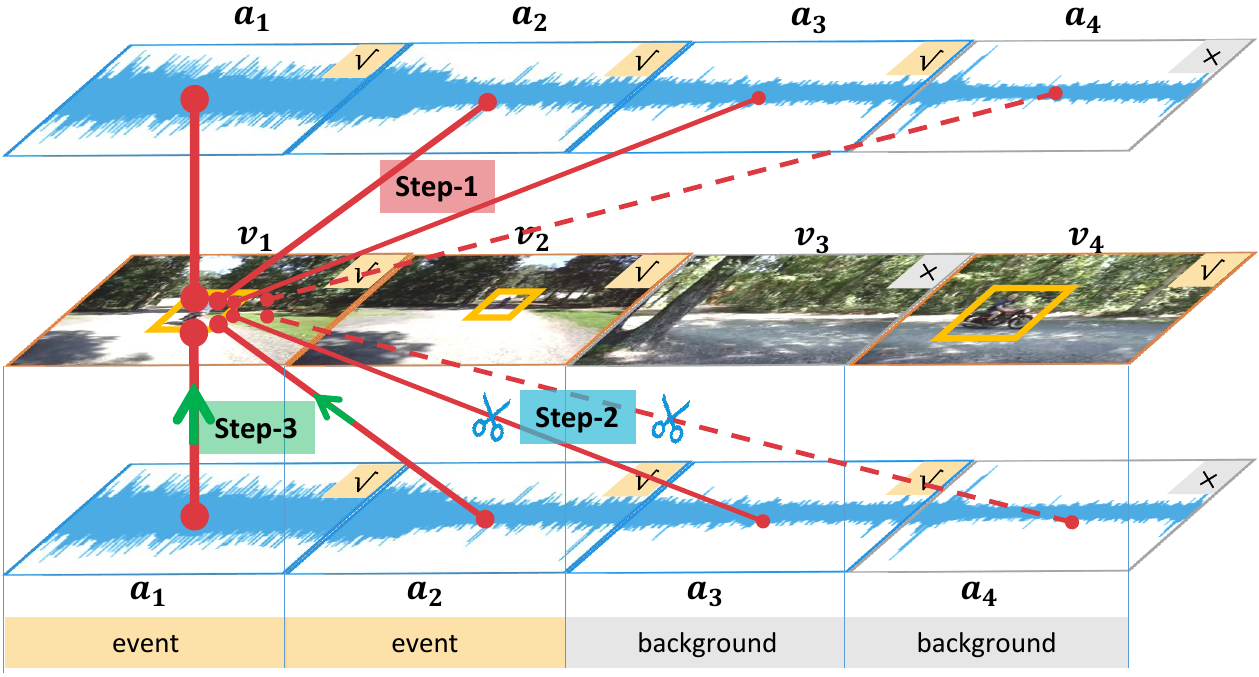}
   \end{center}
   \vspace{-0.2cm}
      \caption{An illustration of the proposed PSP.
      In this example, only the first two video segments contain an audio-visual event, \ie, motorcycle.
      ``$\surd$'' denotes the audio or visual segment describes the event, while ``$\times$'' means not.
      The red lines denote connections of audio-visual pairs,
      solid lines represent connections formed by relevant pairs,
      while dotted lines denote irrelevant pairs.
      The thickness of line reflects the similarity of the audio-visual pair.
      $v_1 \leftrightarrow a_4$ is a {\em negative} connection, formed by irrelevant audio-visual pair with negative similarity value.
      $v_1 \leftrightarrow a_3$ and $v_1 \leftrightarrow a_1/a_2$ are {\em weak} and {\em positive connections} respectively,
      determined via similarity.
      The upper part corresponds to ``Step-1''
       (all-pair connection construction),
      while the lower part denotes ``Step-2''
       (prune the negative and weak connections),
      and the green arrow indicates the direction of feature propagation (``Step-3'').
      }
   \label{figure_2}
   \end{figure}

In \emph{all-pair connection construction},
all the audio-visual pairs are connected.
As shown in Fig.~\ref{figure_2},
here we only display the connections of one visual segment for simplicity,
{\em i.e.}, $\langle v_1 \leftrightarrow a_1/a_2/a_3/a_4 \rangle$.
The strength of these connections are measured
by the similarity between the audio-visual components $\langle{\bm{a}^{\text{lstm}}, \bm{v}^{\text{lstm}}} \rangle$,
computed by,
\begin{equation}
   \bm{\beta}^{\text{va}} = \frac{(\bm{v}^{\text{lstm}}{\bm{W}_1^v})(\bm{a}^{\text{lstm}}{\bm{W}_1^a})^{\top}} {\sqrt{d_l}}, \quad
   \bm{\beta}^{\text{av}} = (\bm{\beta}^{\text{va}})^{\top},
\end{equation}
where $\bm{W}_1^v \mbox{ and } \bm{W}_1^a \in \mathbb{R}^{d_l \times d_h}$
are learnable parameters of linear transformations,
implemented by a linear layer, and
$d_l$ is the dimension of the audio or visual feature. 
$\bm{\beta}^{\text{va}} \mbox{ and } \bm{\beta}^{\text{av}} \in \mathbb{R}^{T \times T}$ are the similarity matrices.

Second, we \emph{prune the negative and weak connections}.
Specifically, the connections constructed in the first step
are divided into three groups according to the similarity values: negative, weak, and positive.
As a classification task, the success of AVE localization highly depends on the richness and correctness of training samples for each class. That is, we aim to collect possibly many and relevant \emph{positive} connections. 
We achieve this goal by filtering out the weak and negative ones, \eg, $v_1 \leftrightarrow a_3$ and $v_1 \leftrightarrow a_4$ as shown in Fig.~\ref{figure_2}.
We begin with processing all the audio-visual pairs
with the ReLU activation function,
cutting off connections with negative similarity values.
Row-wise $\ell_1$ normalization is then performed,
yielding the normalized similarity matrices
$\bm{\beta}^{\text{va}}$ and $\bm{\beta}^{\text{va}}$.

The negative and weak connections are presumably featured by smaller similarity values, so we simply adapt a thresholding method, written as, 
\begin{equation}
\begin{split}
   \bm{\gamma}^{\text{va}} = \bm{\beta}^{\text{va}} \mathbb{I}(\bm{\beta}^{\text{va}}-\tau), \\
   \bm{\gamma}^{\text{av}} = \bm{\beta}^{\text{av}} \mathbb{I}(\bm{\beta}^{\text{av}}-\tau),
   \label{threshold_operation_1}
\end{split}
\end{equation}
where $\tau$ is the hyper-parameter, controlling how many connections will be pruned. $\mathbb{I(\cdot)}$ is an indicator function, which outputs $1$ when the input
is greater than or equal to $0$, and otherwise outputs $0$,
$\bm{\gamma}^{\text{va}}$. After thresholding, row-wise $\ell_1$ normalization is again performed to obtain the final similarity matrices $\bm{\gamma}^{\text{va}},$ $\bm{\gamma}^{\text{av}}\in \mathbb{R}^{T \times T}$.


\emph{Online feature aggregation}. The above step identifies audio (visual) components with high similarities with a given visual (audio) component, \eg, $v_1 \leftrightarrow a_1$ and $v_1 \leftrightarrow a_2$ shown in Fig.~\ref{figure_2}. This is essentially a positive sample propagation process that can be utilized to update the features of audio or visual components. 
Particularly, given the connection weights $\bm{\gamma}^{\text{av}}$ and $\bm{\gamma}^{\text{va}}$, the audio and visual features $\bm{a}^{\text{psp}}$ and $\bm{v}^{\text{psp}}$  are respectively updated as,

\begin{equation}
   \begin{split}
       \bm{a}^{\text{psp}} = \overbrace {\bm{\gamma}^{\text{av}}(\bm{v}^{\text{lstm}}{\bm{W}_2^v})}^{\bm{v}^{\text{pos}}} + \bm{a}^{\text{lstm}},\\
       \bm{v}^{\text{psp}} = \overbrace {\bm{\gamma}^{\text{va}}(\bm{a}^{\text{lstm}}{\bm{W}_2^a})}^{\bm{a}^{\text{pos}}} + \bm{v}^{\text{lstm}},
   \end{split}
   \label{eq:3}
\end{equation}
where $\bm{W}_2^{a}, \bm{W}_2^{v} \in \mathbb{R}^{d_l \times d_l}$ are parameters defining linear transformations, and
$\bm{a}^{\text{psp}}, \bm{v}^{\text{psp}}  \in \mathbb{R}^{T \times d_l}$.

\vspace{-0.08cm}
Generally, the audio (visual) feature $\bm{a}^{\text{psp}}$ ($\bm{v}^{\text{psp}}$) is enhanced by the propagated positive support from the other modality. 
This practice allows us to learn more discriminative audio-visual representations, displayed in Fig.~\ref{figure_5}. More discussions are provided in Sec. \ref{sec:discussion}.

\vspace{-1mm}
\subsection{Classification}\label{sec:classification}
\vspace{-0.1cm}
Before classifier prediction, we transform the visual and audio features into the same embedding space through another linear layer,
and then combine the output through simple averaging, yielding the fusion feature denoted as $\bm{f}^{v \leftrightarrow a}$. This process is written as,
\begin{equation}
   \bm{f}^{v \leftrightarrow a} = \frac {1}{2}[{\mathcal{N}(\bm{v}^{\text{psp}}{\bm{W}_3^v}) + \mathcal{N}({\bm{a}^{\text{psp}}}{\bm{W}_3^a})}],
\end{equation}
where $ \mathcal{N}(\cdot) $ represents layer normalization,
$ \bm{W}_3^v, \bm{W}_3^a \in \mathbb{R}^{d_l \times d_l} $ represent learnable parameters in the linear layers, and $\bm{f}^{v\leftrightarrow a}\in \mathbb{R}^{T \times d_l} $.

For the fully-supervised setting, as shown in Fig.~\ref{figure_3}, the fusion feature is further processed by two FC layers.
The classifier prediction $\bm{o}^{\text{fully}}\in \mathbb{R}^{T \times C}$ can be obtained through a softmax function.

For the weakly supervised setting, different from existing methods~\cite{lin2019dual, tian2018audio, xuan2020cross},
we add a weighting branch on the fully supervised classification module (Fig.~\ref{figure_3}). It is essentially another FC layer that enables the model to further capture the differences between synchronized audio-visual pairs
by dynamically focusing on different event categories.
This process is summarized below,
\begin{equation}
\begin{split}
   \bm{f}^h &= {\bm{f}^{v\leftrightarrow a} \bm{W}_4^{\text{weak}} \bm{W}_5^{\text{weak}}}, \\
   \bm{\phi} &= \sigma(\bm{f}^h \bm{W}_6^{\text{weak}} ), \\
   \bm{o}^{\text{weak}} &= s(f_{\text{avg}}(\bm{f}^h \odot \bm{\Phi})),
\end{split}
\label{eq:o^weak}
\end{equation}
where $\bm{W}_4^{\text{weak}} \in \mathbb{R}^{d_l \times d_h}$, $\bm{W}_5^{\text{weak}} \in \mathbb{R}^{d_h \times C}$,
$\bm{W}_6^{\text{weak}} \in \mathbb{R}^{C\times 1} $ are learnable parameters in the FC layers, and $\bm{f}^h \in \mathbb{R}^{T \times C}$.
$\sigma$ and $s$ denote the sigmoid and softmax operators, respectively.
$\bm{\phi} \in \mathbb{R}^{T \times 1}$ weighs the importance of the temporal video segments, and $\bm{\Phi} \in \mathbb{R}^{T\times C}$ is obtained by duplicating $\bm{\phi}$ for
$C$ times. $\odot$ is the element-wise multiplication,
$f_{\text{avg}}$ is the average operation along the temporal dimension.
The final prediction $\bm{o}^{\text{weak}} \in \mathbb{R}^{1 \times C}$.

For comparison, we denote predictions through a network without the weighting branch as,
\begin{equation}
   \bm{o}_{\text{wo}}^{\text{weak}} = s(f_{\text{avg}}(\bm{f}^h)).
\label{eq:o_wo^weak}
\end{equation}

\vspace{-0.1cm}
\subsection{Objective function}\label{loss_functions}
\vspace{-0.1cm}
\textbf{Fully supervised setting.} Given the network output $\mathbf{o}^{\text{fully}}$ and ground truth $\mathbf{Y}^{\text{fully}}$, we adapt the cross entropy (CE) loss
as the objective function, written as,
\begin{equation}
   \mathcal{L}_{ce} = -\frac {1}{TC}  \sum_{t=1}^{T} \sum_{c=1}^{C} \bm{Y}_{tc}^{fully} {\log(\bm{O}_{tc}^{fully})}
   \label{eq:softmax}
\end{equation}
Recall that each row of $\bm{Y}^{\text{fully}}$ contains a one-hot event label vector, describing the category of the corresponding segment (synchronized audio-visual pair).
As such, this classification loss allows the network to predict which \emph{event category} a video segment contains.

Apart from the CE loss, we propose a new loss item, named audio-visual pair similarity loss $\mathcal{L}_{\text{avps}}$. In principle, it asks the network to produce similar features for a pair of audio and visual components if the pair \emph{contains an event}.
Specifically, for a video composed of $T$ segments, we define label vector $\bm{G} = { \{g_t | g_t \in \{ 0, 1\}, t=1,2,...,T\} \in \mathbb{R}^{1 \times T}} $,
where $g_t$ represents whether the $t^{th}$ segment is an event or background. Next, $\ell_1$ normalization is performed on $\bm{G}$.
We then compute the $\ell_1$ normalized similarity vector $\bm{S} \in \mathbb{R}^{1 \times T}$ between the visual and audio features
\begin{equation}
   \begin{split}
      \bm{S} &= \frac {\bm{v}^{\text{psp}} \odot \bm{a}^{\text{psp}}} {\left \|  \bm{v}^{\text{psp}} \odot \bm{a}^{\text{psp}} \right\|_1},
   \end{split}
   \label{eq:similarity}
\end{equation}
where $\|\cdot\|_1$ calculate the $\ell_1$ norm of a vector. The proposed loss $\mathcal{L}_{\text{avps}}$ is then written as,
\begin{equation}
\begin{split}
   \mathcal{L}_{\text{avps}} &= \mathcal{L}_{\text{MSE}}(\bm{S}, \bm{G}),
\end{split}
\label{eq:avps}
\end{equation}
where $\mathcal{L}_{\text{MSE}}(\cdot, \cdot)$ computes the mean squares error between two vectors.

Combining Eq. \ref{eq:avps} and Eq. \ref{eq:softmax}, the overall objective function for fully-supervised setting $\mathcal{L}_{\text{fully}}$ can be computed by:
\begin{equation}
    \mathcal{L}_{\text{fully}} = \mathcal{L}_{\text{ce}} + \lambda{\mathcal{L}_{\text{avps}}},
   \label{fully_loss}
\end{equation}
where $\lambda$ is a hyper-parameter to balance the two losses.

\textbf{Weakly supervised setting.} For this setting,
following the practice in ~\cite{lin2019dual, xu2020MM}, we adapt the binary cross entropy (BCE) loss, formulated as, 
\begin{equation}
   \begin{split}
   \mathcal{L}_{\text{w-bce}} = \mathcal{L}_{\text{BCE}}(\bm{o}^{\text{weak}}, \bm{Y}^{\text{weak}}), \\
   \mathcal{L}_{\text{wo-bce}} = \mathcal{L}_{\text{BCE}}(\bm{o}_{\text{wo}}^{\text{weak}}, \bm{Y}^{\text{weak}}),
   \end{split}
   \label{eq:weak_bce_loss}
\end{equation}
where $\mathcal{L}_{\text{w-bce}}$ and $\mathcal{L}_{\text{wo-bce}}$ are calculated between the ground-truths
and predictions.
$\bm{o}^{\text{weak}}$ and $\bm{o}_{\text{wo}}^{\text{weak}}$ are predictions obtained with or without the weighting branch (Sec. \ref{sec:classification} and Fig. \ref{figure_3}), respectively.

\subsection{Discussion}\label{sec:discussion}
\vspace{-0.1cm}

\textbf{Detailed examination and meanings of $\bm{v}^{\text{pos}}$ and $\bm{a}^{\text{pos}}$.}
The computation of $\bm{v}^{\text{pos}}$ ($\bm{a}^{\text{pos}}$) is shown in Eq. \ref{eq:3}.
Take $\bm{v}^{\text{pos}}$ for example. The $i^{th}$ row $\bm{v}_{i}^{\text{pos}}$ is the weighted sum of the visual feature $\bm{v}_j^{\text{lstm}} (j=1,2,...,T)$ after linear transformation.
Here the weight, denoted as $\bm{\gamma}_{i}^{\text{av}}$, is exactly the similarity between the audio feature $\bm{a}_i$ and features of all the visual components.
Note that some elements of $\bm{\gamma}_{i}^{\text{av}}$ are zeros since the negative and weak connections are pruned during PSP, so $\bm{v}_i^{\text{pos}}$ is the aggregation result of  those {\em positive} visual features which 
are most relevant to $\bm{a}_i$.

\textbf{Physical meanings of $\bm{v}^{\text{psp}}$ and $\bm{a}^{\text{psp}}$.}
Take $\bm{a}^{\text{psp}}$ for example. From Eq. \ref{eq:3}, we find $\bm{a}^{\text{psp}}$ is composed of two features: the original audio feature $\bm{a}^{\text{lstm}}$ and the aggregation of positive visual features $\bm{v}^{\text{pos}}$.
As discussed above, those positive visual features have large audio-visual similarity values, \ie, small vector angles and similar vector directions. 
Therefore, after being added to $\bm{v}^{\text{pos}}$, the magnitude and direction of vectors representing original audio feature $\bm{a}^{\text{lstm}}$ will be changed to reflect that during training. Such an adjustment in the distribution of audio representation can be verified by the visualization results in Fig. \ref{figure_5}.

\textbf{Why an additional FC layer in the weakly supervised setting?} When fully supervised, clear supervision is known for each segment. 
For the weakly supervised setting, both the ground truth label $\bm{Y}^{\text{weak}} \in \mathbb{R}^{1 \times C}$ and the prediction $\bm{o}^{\text{weak}}\in \mathbb{R}^{1 \times C}$ are obtained through an average pooling operation along the temporal dimension.
Without knowing the supervision for each segment, the baseline approach considers all temporal video segments to have similar weights when calculating the loss.
It makes it harder for the model to focus on video segments that contain an event.
In our design, through the sigmoid activation function, 
we obtain the weights of temporal video segments.
As such, our model can better distinguish these temporal sequences and thus help locate which segments contain an event.

\textbf{Implications of Eq. \ref{eq:avps}.}
As shown in Eq. \ref{eq:softmax}, the classification loss $\mathcal{L}_{\text{ce}}$ prompts the model to correctly predict the event \emph{categories}.
In comparison, $\mathcal{L}_{\text{avps}}$ allows the network to be aware of \emph{whether an event exists in an audio-visual pair}. 
Specifically, if $g_t$ is equal to $1$, the synchronized audio-visual feature should have a higher similarity, and otherwise lower.
Therefore, for an audio (visual) component, $\mathcal{L}_{\text{avps}}$ provides another auxiliary constraint so that the model can better select the most relevant visual (audio) components for feature aggregation during PSP.
Note that $\mathcal{L}_{\text{avps}}$ cannot be adapted in the weakly supervised setting, where the label $g_t$ of each segment is unknown.

\section{Experiment}
\subsection{Experimental setup}
\textbf{Dataset.} Following existing works \cite{lin2019dual, tian2018audio, xu2020MM, xuan2020cross}, we use the AVE dataset~\cite{tian2018audio} that is publicly available. This dataset contains 4,143 videos,
which cover various real-life scenes and can be divided into 28 event categories, \eg, church bell, male speech, acoustic guitar, and dog barking.
Each video sample is evenly partitioned into 10 segments, and each segment is one-second long.
The audio-visual event boundary on the segment level and the event category on the video level are provided.

\textbf{Evaluation metric.} The category label of each segment is predicted in both fully and weakly
supervised settings. Following~\cite{lin2019dual, tian2018audio,xu2020MM,xuan2020cross}, we adapt
the classification accuracy of each segment as the evaluation metric.

\textbf{Implementation details.} We use VGG-19~\cite{simonyan2014very} pretrained
on ImageNet~\cite{krizhevsky2017imagenet} to extract the visual features.
Specifically, 16 frames are sampled from each one-second segment.
We extract the visual feature maps from each frame and use the average map as 
the visual feature for this segment.
For audio features, we first process the raw audio into log-mel
spectrograms and then extract the acoustic features using a VGG-like network~\cite{hershey2017cnn}
pretrained on AudioSet~\cite{gemmeke2017audio}.
Besides, dropout technique is used in the linear layers (Fig. \ref{figure_3}).
Weight $\lambda$ in Eq. \ref{fully_loss} is empirically set to 100.

\subsection{Quantitative Analysis}

\begin{figure*}[t]
   \begin{center}
      \includegraphics[width=\textwidth]{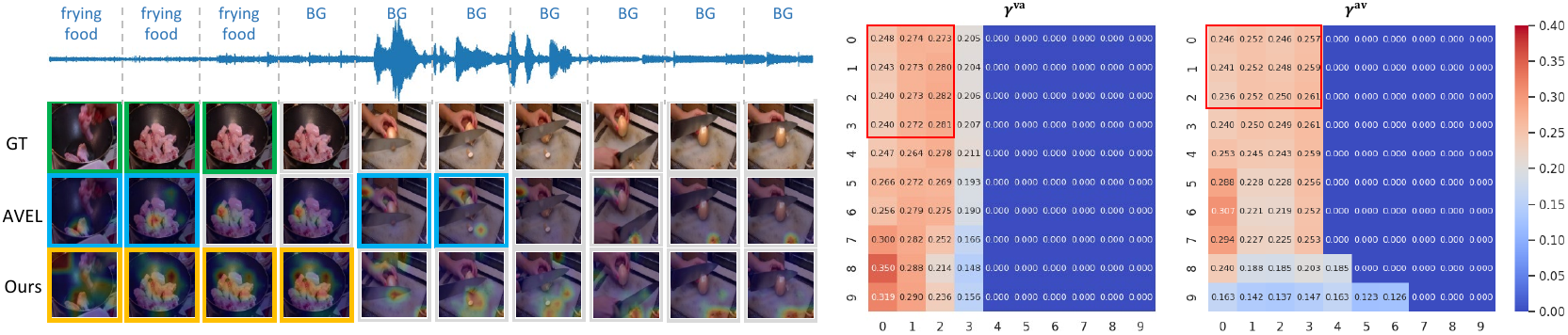}
   \end{center}
      \vspace{-0.3cm}
      \caption{A qualitative example of AVE localization.
      For the video on the \textbf{left}, only the first three segments contain the visual and audio signals of the event \emph{frying food}.
      The green boxes represent ground truth labels.
      The blue and orange boxes indicate predictions
      of AVEL~\cite{tian2018audio} and our method, respectively. Besides, we visualize the attention effect on the images. It is clear that our method produces more accurate localization and that our attended regions better overlap with the sound sources.
      On the \textbf{right}, we visualize the audio-visual similarity matrices $\bm{\gamma}^{\text{va}}$ and $\bm{\gamma}^{\text{av}}$ (Eq.~\ref{eq:3}) after PSP. For $\bm{\gamma}^{\text{va}}$, the x-axis and y-axis correspond to audio and visual features, respectively, and for $\bm{\gamma}^{\text{av}}$ the order is reversed. The red bounding boxes in $\bm{\gamma}^{\text{va}}$ show that the first three audio components are highly correlated with the first four visual components. Besides, negative and weak connections are cut off to 0 in $\bm{\gamma}^{\text{va}}$ and $\bm{\gamma}^{\text{av}}$. The color bar corresponds to the similarity strength, with red denoting high similarities and blue for low similarities.
      }
   \label{figure_8}
  \vspace{-0.3cm}
\end{figure*}

\textbf{The effectiveness of the PSP encoding} can be verified through comparing with an ablation study, \ie, removing it from the localization network (Fig.~\ref{figure_3}). In Table \ref{table_2}, We denote the method without PSP as ``w/o PSP''.
We observe from the table that
the performance drops in both the fully supervised and weakly supervised setting significantly. Specifically, the accuracy decrease is 4.1\% (from 77.8\% to 73.7\%) and 3.3\% (from 73.5\% to 70.2\%) for the two settings, respectively. This experiment clearly validates PSP.

\textbf{Comparison with alternative positive sample selection methods.} In our method, we emphasize that weak and negative samples are filtered out. Here, we compare this strategy with two variants: 1) all connections are used; 2) only negative ones are removed. Results are shown in Table \ref{table_2}. We have two main observations.

First, when all samples are propagated (denoted as ``ASP''), the accuracy drops by 1.9\% and 2.3\% on the two settings, respectively. This shows that it is essential to have a selection process before feature aggregation instead of utilizing all the connections. In fact, the ASP variant shares the same spirit with HAN~\cite{tian2020han}.

Second, when we only remove the negative connections (\ie, those with a similarity value below $\tau = 0$), the system is inferior to the full method. Specifically, the classification accuracy decreases by 1.8\% and 2.3\% under the two settings, which validates the effectiveness of filtering out the negative connections. 


\begin{table}[t]\small
   \setlength{\tabcolsep}{3.5mm}
   \begin{center}
   \begin{tabular}{|l|c|c|}
   \hline
   Method & Fully-supervised & Weakly-supervised \\
   \hline
   w/o PSP & 73.7 & 70.2 \\
   ASP  &  75.9 & 71.2\\
   WPSP & 76.0 & 71.2 \\
   SAPSP & 75.4 & 70.8 \\\hline
   PSP (ours) & {\bf 77.8} & {\bf 73.5} \\
   \hline
   \end{tabular}
   \end{center}
   \vspace{-0.1cm}
   \caption{Ablation studies of the proposed PSP, measured by accuracy(\%) on the AVE dataset. ``w/o'' denotes ``without''. ``ASP'' means retaining all connections ($\tau = -\infty$), while ``WPSP'' uses the weak and positive ones ($\tau = 0$). ``SAPSP'' represents adding self-attention to the feature extractor.}
   \label{table_2}
\end{table}

\textbf{Comparison with adding self-attention \cite{vaswani2017attention} to the feature extractor.}
Self-attention~\cite{vaswani2017attention} is widely used in existing methods
~\cite{tian2020han, wu2019dual, xu2020MM, xuan2020cross} to capture relationships within single modality.
To explore whether it is useful in our system, we add a self-attention module before the Bi-LSTMs and denote it as the ``SAPSP'' method.
As shown in Table \ref{table_2}, the performance surprisingly decreases by
2.4\% and 2.7\%
under fully and weakly supervised settings, respectively.
We speculate that the PSP module is sufficient to describe the cross-modality
while implicitly reveals the intra-modality correlations. For example, in PSP a visual component is constrained to have similar features with multiple audio components describing the same event. Such cross-modality similarity at the same time implies that the similarity of the involved audio components to be high. In our future work, we will study in-depth the intra-modality and inter-modality similarities.

\textbf{Benefit of the audio-visual pair similarity loss $\mathcal{L}_{\text{avps}}$.}
We respectively adapt $\mathcal{L}_{\text{ce}}$
and $\mathcal{L}_{\text{ce}}+\lambda\mathcal{L}_{\text{avps}}$ as the objective function for model training. Two baselines are used: our PSP system and the AVEL system \cite{tian2018audio}. Results are presented in Table \ref{table_1}. 
We can clearly see that $\mathcal{L}_{\text{avps}}$ improves the accuracy when the system is fully supervised. The improvement is 1.2\% and 1.5\% for PSP and AVEL, respectively. These results verify the role of $\mathcal{L}_{\text{avps}}$ as an auxiliary restriction to help to select the positive audio-visual pairs for feature aggregation.

%


\textbf{Improvement from the additional FC in the weakly supervised setting.}
In the weakly supervised setting, the major difference between our classification module and traditional methods~\cite{lin2019dual, tian2018audio, xuan2020cross} consists in the weighting branch (Fig. \ref{figure_3}).
To evaluate its effectiveness, we also implement this branch on top of the PSP and AVEL baselines.
The results are shown in the last two rows of Table \ref{table_1}. We find that the performance
of PSP and AVEL is improved by
1.9\% and 2.3\%,
respectively.
We argue the additional weighting branch within the designed classification module allows the model to give different weights to the temporal sequences, thus benefiting the localization of the target video segments. These results confirm the effectiveness of the proposed improvements. We refer readers to Sec. \ref{sec:discussion} for discussions on the two techniques.


\textbf{Sensitivity to hyper-parameter $\tau$.} The selection process is controlled by $\tau$, determining how many connections will be cut off.
Its influence on the system accuracy is shown in Table~\ref{table_3}. We observe that overall the accuracy remains stable when $\tau$ varies between 0 and 0.115 and that the highest accuracy is achieved when $\tau=0.095$. 
For different videos, the proportion of segments that are cut off highly depends on the video itself. If the whole video contains the same event of interest, it is likely that most will be retained in training; if a video contains lots of background, the same threshold will cut off more of its content.

\begin{table}[t]\small
   \setlength{\tabcolsep}{2.5mm}
   \begin{center}
   \begin{tabular}{|l|l|c|c|}
   \hline
   \multicolumn{1}{|l|}{Setting} & Method     & PSP (ours) & AVEL~\cite{tian2018audio} \\ \hline
   \multirow{2}{*}{fully}        & $\mathcal{L}_{ce}$       & 76.6     & 69.8*   \\
                                 & $\mathcal{L}_{\text{ce}} + \lambda\mathcal{L}_{\text{avps}}$ & {\bf 77.8}               & \bf 71.3*   \\ \hline
   \multirow{2}{*}{weakly}       & w/o weight. branch  & 71.6              & 66.9*   \\
                                 & w/ weight. branch   & {\bf 73.5}        & \bf 69.2*   \\ \hline
   \end{tabular}
   \end{center}
   \vspace{-0.1cm}
   \caption{Method comparison on the AVE dataset under two settings. We evaluate 1) the audio-visual pair similarity loss $\mathcal{L}_{\text{avps}}$ under the fully supervised setting, and 2) the weighting branch under the weakly supervised setting. The two improvements are implemented on top of our system and AVEL \cite{tian2018audio}. Under AVEL, * denotes that the number is produced by us. We use \textbf{bold} font to show the higher performance brought by our technique.}
   \label{table_1}
\end{table}

\begin{table}[t]
  \vspace{-0.1cm}
   \setlength{\tabcolsep}{1.95mm}
   \small
   \begin{center}
   \begin{tabular}{|l|c|c|c|c|c|}
   \hline
   $\tau$ & 0 & 0.025 & 0.075 & 0.095 & 0.115 \\
   \hline
   Fully-supervised & 75.9 & 76.1 & 75.3 & {\bf 77.8} & 76.6 \\
   Weakly-supervised & 71.2 & 71.7 & 70.4 & {\bf 73.5} & 72.8  \\
   \hline
   \end{tabular}
   \end{center}
   \vspace{-0.2cm}
   \caption{Impact of various values of $\tau$ on the system accuracy. Results on the two setting are shown.}
   \label{table_3}
\end{table}


\begin{figure*}[t]
   \setlength{\abovecaptionskip}{0.cm}
   \vspace{-1mm}
   \includegraphics[width=\textwidth]{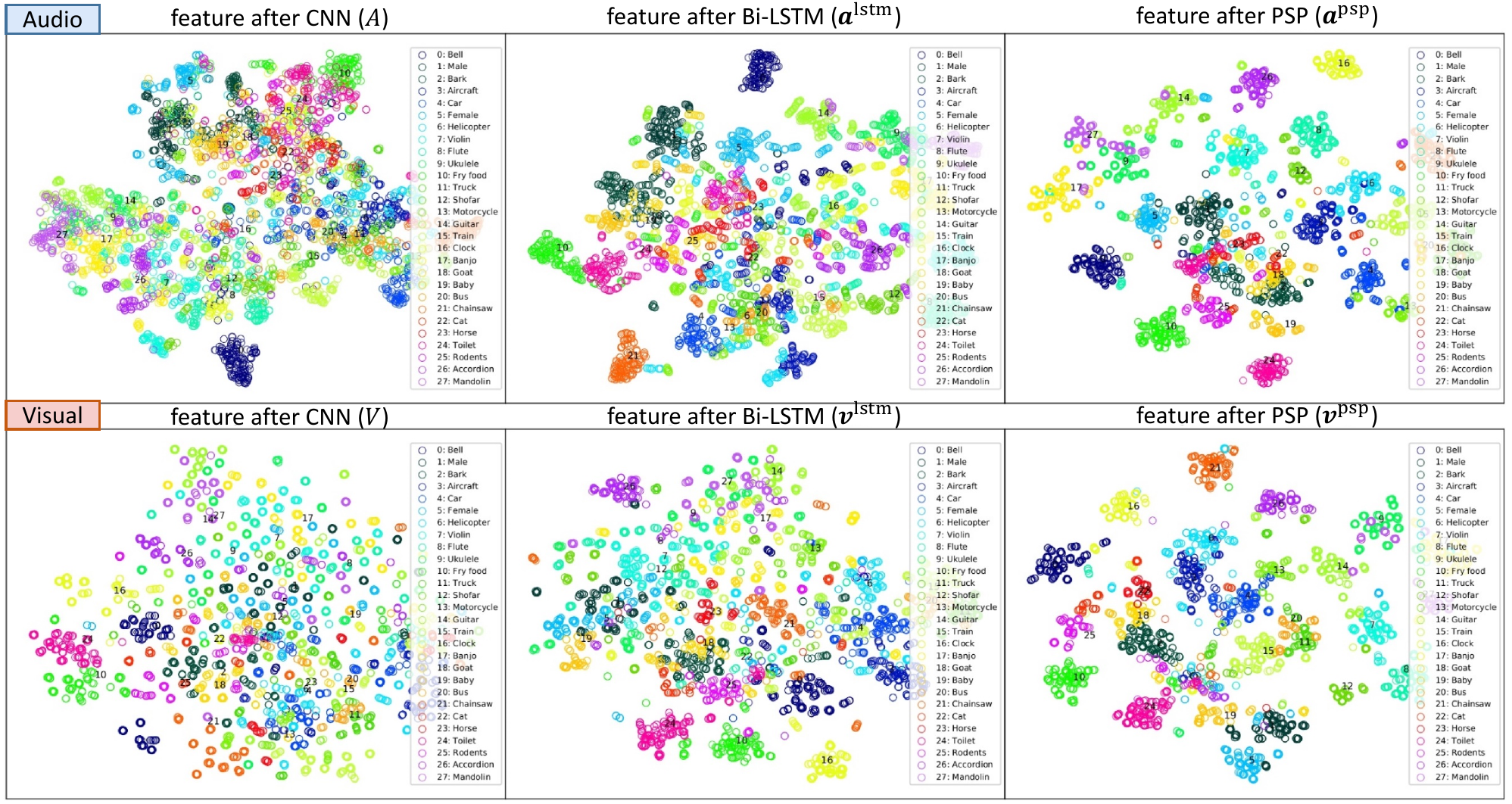}
   \vspace{-3mm}
   \caption{TSNE \cite{maaten2008visualizing} visualization of audio and visual feature distributions under the fully-supervised setting.
   The data all come from the validation set.
   (\textbf{Row 1:}) audio features. (\textbf{Row 2:}) visual features.
   (\textbf{Column 1:}) the CNN features. (\textbf{Column 2:}) features after Bi-LSTM encoding. (\textbf{Column 3:}) features after PSP encoding. 
   We observe that features after PSP are much better clustered into individual classes than the Bi-LSTM and CNN features. Different colors represent different classes. 
   Best view in color and zoom in.
   }
   \setlength{\belowdisplayskip}{1pt}
   \label{figure_5}
  \vspace{-0.2cm}
\end{figure*}

\textbf{Comparison with the state of the art.} We compare our method with the state of the art in Table~\ref{table_4}, where we report superior results: \textbf{the classification accuracy is 77.8\% and 73.5\% for the fully and weakly supervised settings, respectively.}
Compared with the baseline feature extractor AVEL~\cite{tian2018audio}, we exceed it by 9.2\% and 6.8\% under the fully and weakly supervised settings, respectively.
This can also be proved by the results shown in Table \ref{table_1} where the numbers of AVEL are reproduced by ourselves.
Moreover, while the AVGA module~\cite{tian2018audio} adapted in our system is slightly lower (0.6\%) than the recent audio-guided spatial-channel attention (AGSCA)~\cite{xu2020MM}, our overall system manages to obtain higher accuracy than AGSCA. This can be attributed to both the PSP and our system design.

\begin{table}[t]\small
   \begin{center}
   \setlength{\tabcolsep}{3mm}
   \begin{tabular}{|l|c|c|}
   \hline
   Method  & Fully-supervised   & Weakly-supervised \\ \hline
   AVEL~\cite{tian2018audio}      & 68.6   & 66.7 \\
   AVSDN~\cite{lin2019dual}       & 72.6   & 67.3 \\
   CMAN~\cite{xuan2020cross}      & 73.3*   & 70.4* \\
   DAM~\cite{wu2019dual}          & 74.5   & -    \\
   AVRB~\cite{ramaswamy2020see}   & 74.8   & 68.9 \\
   AVIN~\cite{ramaswamy2020makes} & 75.2   & 69.4 \\
   AVT~\cite{Lin_2020_ACCV}       & 76.8   & 70.2 \\
   CMRA~\cite{xu2020MM}           & 77.4   & 72.9 \\ \hline
   PSP (Ours)                      & {\bf 77.8}   & {\bf 73.5} \\ \hline
   \end{tabular}
   \end{center}
   \vspace{-0.1cm}
   \caption{Comparison with the state-of-the-art methods under two settings, measured by accuracy(\%) on the AVE dataset. * indicates the number is reproduced by us.}
   \label{table_4}
\end{table}


\subsection{Qualitative analysis}\label{visualization}
We start by presenting an example of audio-visual event localization in Fig.~\ref{figure_8}.
The event in this sample is difficult to predict because the visual images are changeable
and the audio signals are mixed with background noise.
1). While both our method and AVEL \cite{tian2018audio} use the AVGA attention, we show that our method enables better attention to visual regions closely related to sound sources.
As displayed in Fig.~\ref{figure_8}, for the event of {\em frying food},
our attended regions include both the frying chicken thighs and the pot, especially in the first four segments.
In comparison, AVEL only finds the thighs and very small receptive fields.
2). Our method has a better prediction result.
AVEL seems to make decisions merely according to synchronized audio-visual segments while
our method can pay attention to visual and audio components that are at different time stamps. For example,
AVEL incorrectly regards the fifth and sixth segments as the {\em frying food} event, ignoring the third and fourth segments which are more relevant to the event.
3). We visualize the similarity matrices $\bm{\gamma}^{\text{va}}$ and $\bm{\gamma}^{\text{av}}$ in Fig.~\ref{figure_8}.
We find that only a small percentage of all the audio-visual connections are retained after PSP selection
and are closely related to the event. For example,
for the first four visual components describing the target event,
they tend to build strong connections (large similarity values) with the first three audio components containing the sound of the event.
Such a propagation mechanism is critical for AVE localization because 
more discriminative audio-visual features can be identified with these {\em positive} connections and subsequently used in classifier training. 
Through backpropagation, it allows the model to be able to attend to broader and more sound-relevant regions in the visual images.

We then visualize the data distribution of features processed by different stages in our framework using TSNE~\cite{maaten2008visualizing} (Fig. \ref{figure_5}). We first find that the CNN-based audio and visual features are not very well clustered. This is because they are at a relatively low level in the network hierarchy encoding limited semantics. Then, after Bi-LSTM, features of some categories (\eg, \emph{rodents} and \emph{Fry food}) can be better clustered compared with the CNN features, but most are still disordered and highly mixed. Further, after PSP, the features are much better clustered: cohesive within the same class and divergent between different classes. 
This reflects that the audio-visual representations gain stronger discriminative abilities along the pipeline of our method. 

\section{Conclusion}
For the AVE localization problem, we propose a positive sample propagation (PSP) method, which identifies and exploits relevant but unsynchronized audio and visual samples to enrich the encoded features. We find that negative and weak connections, even though having small weights, have a detrimental effect on the system, and thus have to be completely removed. 
Further, for the fully supervised setting, we propose an audio-visual pair similarity loss to supervise feature learning from a complementary way: whether a segment contains an event. 
For the weakly supervised setting, we insert a weighting branch to the classification module inject temporal importance to the features.
Extensive experiments validate the effectiveness of these techniques.

\vspace{0.3cm}
\noindent
{\textbf{Acknowledgement.}
This work was supported by
the National Key Research and Development Program of China (2018YFB0804205),
the National Natural Science Foundation of China (61725203, 61732008).}

\newpage
{\small
\bibliographystyle{ieee_fullname}
\bibliography{egbib}
}

\end{document}